\documentclass{article}
\usepackage{cite}
\usepackage{spconf}
\usepackage{amsmath,amssymb,amsfonts, amscd, amsthm, dsfont, amsopn}
\usepackage{graphicx}
\usepackage{textcomp}
\usepackage{booktabs}
\usepackage{xcolor}
\usepackage{array}
\usepackage{amsthm}
\usepackage{subfigure,enumitem}
\usepackage{comment,hyperref}
\usepackage{algorithm}
\usepackage{algpseudocode}

\def\BibTeX{{\rm B\kern-.05em{\sc i\kern-.025em b}\kern-.08em
    T\kern-.1667em\lower.7ex\hbox{E}\kern-.125emX}}
\usepackage{color}
\usepackage{xcolor}

\allowdisplaybreaks
\DeclareMathOperator*{\argmax}{arg\,max}

\title{Neural Stochastic Differential Equations with Change Points: \\ A Generative Adversarial Approach}

\name{Zhongchang Sun$^{\star\dagger}$ \qquad Yousef El-Laham$^{\star}$ \qquad Svitlana Vyetrenko$^{\star}$}
\address{J.P. Morgan AI Research$^{\star}$, University at Buffalo$^{\dagger}$}

\usepackage{setspace}  

\begin{document}
\ninept
% \onecolumn
% \doublespacing

\maketitle

\begin{abstract}
Stochastic differential equations (SDEs) have been widely used to model real world random phenomena. Existing works mainly focus on the case where the time series is modeled by a single SDE, which might be restrictive for modeling time series with distributional shift. In this work, we propose a change point detection algorithm for time series modeled as neural SDEs. Given a time series dataset, the proposed method jointly learns the unknown change points and the parameters of distinct neural SDE models corresponding to each change point. 
Specifically, the SDEs are learned under the framework of generative adversarial networks (GANs) and the change points are detected based on the output of the GAN discriminator in a forward pass. %At each step of the proposed algorithm, the change points and the SDE model parameters are updated in an alternating fashion. 
Numerical results on both synthetic and real datasets are provided to validate the performance of the algorithm in comparison to classical change point detection benchmarks, standard GAN-based neural SDEs, and other state-of-the-art deep generative models for time series data.
\end{abstract}

\vspace{-0.1cm}
\section{Introduction}
\vspace{-0.1cm}
Stochastic differential equations (SDEs) are a class of mathematical equations %that are usually 
used to model continuous-time stochastic processes \cite{lelievre2016parial, soboleva2003population,huillet2007on}, with applications ranging from finance and physics to biology and engineering. Recently, neural SDEs \cite{tzen2019theoretical,li2020scalable, gierjatowicz2022robust, liu2019neural, song2020score, kidger2021neural, hasan2022} have been proposed as a means to integrate neural networks with SDEs, providing a more flexible approach for modeling %and learning from 
sequential data. In \cite{kidger2021neural}, the authors established a novel connection between neural SDEs and generative adversarial networks (GANs), showing that certain classes of neural SDEs can be interpreted as infinite-dimensional GANs. In \cite{hasan2022}, a variational autoencoder (VAE) framework for identifying latent SDEs from noisy observations was proposed based on the Euler-Maruyama approximation of SDE solutions.  Existing works on neural SDEs mainly focus on the case where the time series is modeled by a single SDE; however, in real-world applications, the underlying dynamics of the data may change over time. For example, financial time series may exhibit sharp distributional shifts due to exogenous factors (e.g., global financial crisis, the COVID-19 pandemic). To train the neural SDEs, it is {common to} %usually 
assume that the drift and diffusion terms are Lipschitz continuous. {This assumption is restrictive, in the sense that a single neural SDE that with Lipschitz smooth drift and diffusion cannot effectively model time series with sudden distributional shifts}. This motivates us to study the change point detection problem of SDEs and model the time series as multiple SDEs conditioned on the change points. 

Change {point} detection \cite{nikiforov1995a, veeravalli2014quickest} is a critical aspect of time series analysis, especially in domains such as finance, climate science, and sensor data processing, where abrupt shifts in behavior can have profound implications. By identifying change points, we can partition the time series into distinct segments where each segment is described by a different SDE model. This adaptation allows us to capture the specific characteristics and uncertainties within each segment, leading to a more precise understanding of the underlying processes. In \cite{iacus2010numerical, kovavrik2013volatility}, SDEs are applied to detect change {points} in time series. However, the drift and diffusion functions in \cite{iacus2010numerical, kovavrik2013volatility} are characterized by a restricted number of parameters instead of neural networks, which constrains the overall model capacity of SDEs. In \cite{ryzhikov2022latent}, latent neural SDEs are introduced to detect changes in time series, where a single SDE in the latent space is assumed and is trained using VAEs. In this work, it is assumed that there is a prior SDE with a known diffusion term in the latent space for the tractability purposes of the loss function. However, this assumption is too restrictive since the training data might not necessarily conform to this latent SDE. Neural jump SDEs (JSDEs) were proposed in \cite{jia2019neural}, which combine temporal point processes and neural ordinary differential equations (ODEs) \cite{chen2018neural} to model both continuous dynamics and abrupt changes. Compared with the neural SDEs, the continuous dynamics of neural JSDEs is deterministic and the randomness only comes from the temporal point process. Similarly,  stochastic deep latent state space model \cite{rubanova2019latent, gu2021efficiently} combined with ODE-based model are introduced in \cite{zhou2023deep} to increase the modeling capacity of ODEs. However, a prior on the latent variable sequence is needed and no change detection is involved in this method.

In this work, we develop a novel approach for modeling change points in neural SDEs based on the GAN framework presented in \cite{kidger2021neural}, which enhances the expressive capacity of neural SDEs. %. Since GANs are renowned for their ability to learn and generate complex and high-dimensional data distributions, training SDEs as GANs \cite{kidger2021neural} can effectively learn the underlying data distribution. 
%Specifically, we propose to jointly identify the change point and learn the underlying SDE models, {where} at each step {of the algorithm}, we alternate between updating the parameters of the GAN components by running stochastic gradient descent and detecting the change points using the learned GAN discriminator. 
Our specific contributions are as follows:
{\begin{itemize}
  \item We propose a framework and training algorithm for modeling neural SDEs with change points. The proposed algorithm alternates between  detecting change points (while holding the model parameters fixed) and optimizing the GAN parameters (while holding the change points fixed). 
  % \item We propose to alternatively update the change points and SDE model parameters to ensure an efficient convergence.
  \item We propose a change point detection scheme for neural SDEs (trained as GANs) by leveraging the learned GAN discriminator as a means to approximate the Wasserstein distance between time series samples. Specifically, we first partition the training data into multiple segments based on the sliding window approach and then input them sequentially into the GAN discriminator to get a sequence of scores. The change point estimate is then updated by specifying the change point of the score sequence, at which the approximated Wasserstein distance between two consecutive segments is the largest.
  \item We demonstrate the effectiveness and versatility of our approach through extensive experiments on synthetic and real-world datasets.
  \end{itemize}}

\section{Problem Formulation}\label{sec:formulation}
{We consider SDEs of the following form:}
\begin{flalign}\label{eq:sde}
dX_t = f(t, X_t)dt + g(t, X_t)\circ dW_t,
\end{flalign}
where $X_0 \sim \mu$ is the initial state following the initial distribution $\mu$, $X = \{X_t\}_{t\in[0, T]}$ is a continuous $\mathbb{R}^x$-valued stochastic process, ``$\circ$" denotes that the SDE is understood using Stratonovich integration, $f:[0, T]\times \mathbb{R}^x\rightarrow\mathbb{R}^x$ is called the drift function that describes the deterministic evolution of the stochastic process, $g:[0, T]\times \mathbb{R}^{x}\rightarrow\mathbb{R}^{x\times w}$ is called the diffusion function and $W = \{W_t\}_{t\geq 0}$ is a $w$-dimensional Brownian motion representing the random noise in the sample path. Unlike ODEs, SDEs do not always have unique solutions. {We say that} $X = \{X_t\}_{t\geq 0}$ is a strong solution of the SDE \eqref{eq:sde} if it satisfies \eqref{eq:sde} for each sample path of the Wiener process $\{W_t\}_{t\geq 0}$ and for all $t$ in the defined time interval almost surely.

Due to the large capacity of neural networks for function approximation, %the 
neural SDEs {have been} proposed,  {which model} the drift and diffusion terms via neural networks. When training the neural SDEs, the drift function $f$ and the diffusion function $g$ are assumed to be Lipschitz continuous {so that a unique strong solution to the
SDEs exists \cite{kloeden1992stochastic}}. Therefore, when there are changes in the dynamics of the stochastic process, it is not accurate to model the stochastic process as a single {neural} SDE. In this paper, {we turn to an alternative approach, where we leverage multiple neural SDE models conditioned on change points to model the dynamics of a continuous-time stochastic process.} Our goal is to jointly detect the change of the dynamics in the time series and model the time series with multiple SDEs conditioned on the change points.

\vspace{-0.1cm}
\section{{Background}}\label{sec:background}
\vspace{-0.1cm}
\subsection{{Neural SDEs as GANs}}
In this section, we show that fitting the SDEs can be approached using WGANs \cite{kidger2021neural}. {WGANs \cite{arjovsky2017wasserstein} utilize a generator network and a discriminator network, where the loss function is defined using the Wasserstein distance. WGANs enforce Lipschitz continuity on the discriminator through gradient penalties, fostering training stability and convergence while minimizing mode collapse.}

Let $Y_{\text{true}}$ be the ground truth of the SDE trajectory which is a random variable on the path space. Let $V \sim \mathcal{N}(0, I_v)$ be a $v$-dimensional random {Gaussian} noise. 
% Let 
% \begin{flalign}
% &\zeta_\theta:\mathbb{R}^v \rightarrow \mathbb{R}^x,\nonumber\\& \mu_\theta:[0, T]\times\mathbb{R}^x \rightarrow \mathbb{R}^x,\nonumber\\& \sigma_\theta:[0, T]\times \mathbb{R}^{x\times w},\nonumber\\&\alpha_\theta \in\mathbb{R}^{y\times x},\nonumber \\& \beta_\theta\in\mathbb{R}^y,
% \end{flalign}
The generator maps $V$ to a trajectory, which is the solution to the following neural SDE:
\begin{flalign}
X_0 &= \zeta_\theta(V),\nonumber\\ dX_t &= \mu_\theta(t, X_t)dt + \sigma_\theta(t, X_t)\circ dW_t,\nonumber\\ Y_t &= \alpha_\theta X_t + \beta_\theta,
\end{flalign}
where $\zeta_\theta, \mu_\theta$ and $\sigma_\theta$ are (Lipschitz) neural networks and are parameterized by $\theta$. {$\alpha_\theta$ and $\beta_\theta$ are vectors that are jointly optimized.} The generator networks {are optimized} %is trained 
{so} that the generated sample {on path space} $Y_t$ is close to the ground {truth trajectory} $Y_{\text{true}}$.

{For the discriminator, a} neural controlled differential equation (CDE) {is utilized} since it {can} take {an} infinite-dimensional sample path as input and can output a scalar score, {which in practice measures the realism of path with respect to the real data.}
% Let
% \begin{flalign}
% &\xi_\phi:\mathbb{R}^y\rightarrow\mathbb{R}^h,\nonumber\\&
% f_\phi:[0, T]\times\mathbb{R}^h \rightarrow \mathbb{R}^h,\nonumber\\& g_\phi:[0, T]\times \mathbb{R}^h \rightarrow \mathbb{R}^{h\times y},\nonumber\\& m_\phi\in\mathbb{R}^h,
% \end{flalign}
The discriminator has the following form:
\begin{flalign}
H_0 &= \xi_\phi(Y_0),\nonumber\\ dH_t &= f_\phi(t, H_t)dt + g_\phi(t, H_t)\circ dY_t,\nonumber\\ D&=m_\phi\cdot H_T,
\end{flalign}
where $\xi_\phi, f_\phi$ and $g_\phi$ are (Lipschitz) neural networks and are parameterized by $\phi$, $H:[0, T]\rightarrow\mathbb{R}^h$ is the solution to this SDE and $m_\phi$ maps the terminal state $H_T$ to a scalar $D$. 

Let $Y_\theta: (V, \{W\}_{t\geq 0}) \rightarrow Y$ be the overall action of the generator and $D_\phi: Y \rightarrow D$ be the overall action of the discriminator. Let $\mathbf{y}$ be the collection of the training data. The training loss is defined as the Wasserstein GANs, where the generator is trained to minimize 
\begin{flalign}
E_{V, W}[D_\phi(Y_\theta(V, W))],
\end{flalign}
and the discriminator is trained to maximize 
\begin{flalign}\label{eq:loss}
E_{V, W}[D_\phi(Y_\theta(V, W))] - E_{\mathbf{y}}[D_\phi(\hat y)].
\end{flalign}
The goal is to minimize the Wasserstein distance between the true data distribution and the generated distribution \cite{arjovsky2017wasserstein}. The loss functions can be optimized using stochastic optimization techniques (e.g., SGD \cite{lecun1998gradient}, RMSprop \cite{goodfellow2016deep}, and Adam \cite{kingma2014adam}).

\subsection{{Wasserstein Two-Sample Testing}}
For training SDEs as GANs \cite{kidger2021neural}, the training loss can be viewed as the Wasserstein distance between the training samples and the generated samples. Therefore, the learned model can be used to approximate the Wasserstein distance between two time series, which motivates us to design change detection algorithm by leveraging the popular Wasserstein two-sample test \cite{ramdas2017wasserstein}. 
In this section, we provide a brief introduction for the Wasserstein two-sample test. The details of our algorithm are presented in the following section.

The Wasserstein two-sample test \cite{ramdas2017wasserstein} is a statistical method used to compare two sets of data and determine if they originate from the same distribution. Unlike traditional tests that focus on comparing means or variances, the Wasserstein two-sample test computes the Wasserstein distance between the empirical distributions of the samples which measures the minimum amount of cost required to transform one distribution into the other. Specifically, given independent and identically distributed (i.i.d.) samples $X_1, \cdots, X_m \sim P$ and $Y_1, \cdots, Y_n \sim Q$ where $P, Q$ are probability measures on $\mathbb{R}^d$, let $P_m, Q_n$ denote the empirical distributions of $X_1, \cdots, X_m$ and $Y_1, \cdots, Y_n$ respectively. Given an exponent $p\geq 1$, the $p$-Wasserstein distance between $P_m$ and $Q_n$ is defined as
\begin{flalign}
{\cal W}(P_m, Q_n) = \Big(\inf_{\pi\in\Pi(P_m, Q_n)} \int_{\mathbb{R}^d\times\mathbb{R}^d} \|X-Y\|^pd\pi\Big)^{\frac{1}{p}}\nonumber,
\end{flalign}
where $\Pi(P_m, Q_n)$ is the collection of all joint probability distribution on $\mathbb{R}^d\times\mathbb{R}^d$ with marginal distribution $P_m, Q_n$. 

The Wasserstein two-sample test \cite{ramdas2017wasserstein} is particularly useful for high-dimensional data and can provide more informative insights into the dissimilarities between distributions. {The Wasserstein distance has also found other applications in various aspects of statistical inference such as goodness-of-fit testing \cite{hallin2021multivariate} and change detection \cite{faber2021watch}. In \cite{faber2021watch}, Wasserstein barycenters were used to capture changes in distribution.
}

\section{Change {Point} Detection {in Neural SDEs}}\label{sec:algorithm}
In this section, we investigate the problem of modeling change points in neural SDE models based on the GAN framework. To make the presentation more concise, we consider the case where there is one change point and later discuss a straightforward extension to case of multiple change points. Note that since we detect the change point and learn the SDE models in a data-driven manner and the data is not independent over time, it is challenging to directly detect the change using classical change detection algorithms such as the CuSum algorithm \cite{page1954continuous}. Observe that the training loss in $\eqref{eq:loss}$ is defined to approximate the Wasserstein distance between the training data and the generated samples, given the trained discriminator, we can approximate the Wasserstein distance between two time series. Therefore, we propose to detect change by leveraging the idea of Wasserstein two-sample test and alternatively update the parameters for the SDEs and change point estimate.

{\bf Algorithm summary:} Our training algorithm is summarized as follows. Firstly, we initialize the change point estimate $\nu$ and the neural network parameters $\theta_0, \theta_1, \phi$ for the generator and the discriminator. Secondly, based on the change point estimate $\nu$, we partition the training data and run different SDE models for each segment and update the parameters of the GANs. Thirdly, we apply a sliding window method to get multiple segments of the training data and then input them sequentially into the discriminator. As we iterate through the time series, a sequence of scores is returned. The difference of scores between two segments can be viewed as the Wasserstein distance between two segments. Therefore, the change point estimate is then updated by specifying the change point of the score sequence. Figure \ref{fig:diagram} shows a flow diagram summarizing our training algorithm. Specifically, at each step, we alternate between the following two update steps:

\begin{figure}[t]
\centering
\includegraphics[width=3in]{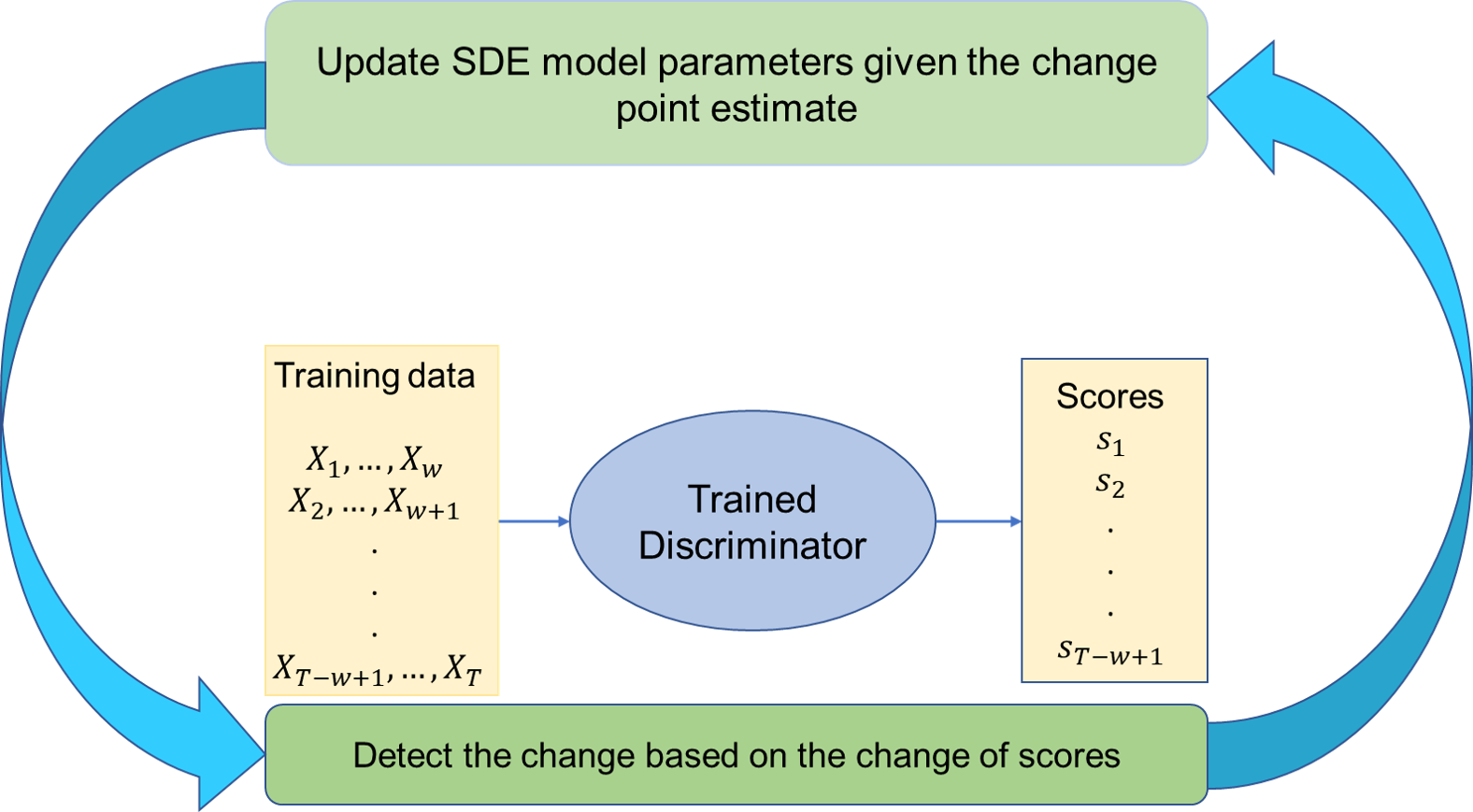}
\caption{Flow diagram of our training algorithm.}\label{fig:diagram}
\end{figure}

{\bf Model parameters update:} Based on the change point estimate $\nu$, we use sample paths $X_{1:\nu-1}$ as training samples to optimize the parameter $\theta_0$ of the neural SDE (before the change happens):
\begin{flalign}
   dX_t = \mu_{\theta_0}(t, X_t)dt + \sigma_{\theta_0}(t, X_t)\circ dW_t, 
\end{flalign}
and use sample paths $X_{\nu:T}$ as training samples to optimize the parameters $\theta_1$ of the neural SDE (after the change happens):
\begin{flalign}
    dX_t = \mu_{\theta_1}(t, X_t)dt + \sigma_{\theta_1}(t, X_t)\circ dW_t.
\end{flalign}
We also update the parameter $\phi$ of the discriminator $D_\phi$ based on the generated trajectory $Y_{1:T}$. 

{\bf Change point update:} After the SDEs model parameters are updated, we update the change point estimate. Consider a sliding window of size $w$. Note that this window size is a hyperparameter of the algorithm that can be tuned in practice. We partition the observed sample path into different segments $X_{1:w}, X_{2:w+1}, \cdots, X_{T-w+1: T}$.
% Note that it's challenging to directly compare the difference between two time series. We further observe that the discriminator is trained to distinguish the real data and generated data and it takes time series as input and output a scalar score. This motivates us to convert the time series to a sequence of scalar scores using the discriminator and design the change point algorithm based on the scores. 
We pass each segment $X_{t:t+w}$ into the discriminator and denote the returned score by $s_t$: 
\begin{flalign}
s_t = D_\phi(X_{t:t+w}), \quad t=1, 2, \ldots, T-w+1.
\end{flalign}
The subsequences $X_{1:w}, X_{2:w+1}, \cdots, X_{T-w+1: T}$ are thus converted to a sequence of scores $s_1, s_2, \cdots, s_{T-w+1}$. We define the average score over all training samples using the arithmetic average:
\begin{flalign}\label{eq:averagescore}
\bar{s}_t = \frac{1}{N}\sum_{i=1}^N D_\phi(X_{t:t+w}^{(i)}).
\end{flalign}
The difference between two average scores can be viewed as the Wasserstein distance between two corresponding segments. 
Sequentially, at each time $t$, we compare the approximated Wasserstein distance between two consecutive segments $\bar{s}_t- \bar{s}_{t-1}$ with a pre-specified threshold $\gamma$ to distinguish between two hypotheses: $\mathcal{H}_0$: the change happens at time $t$; and $\mathcal{H}_1$: the change happens after time $t$. When $\bar{s}_t- \bar{s}_{t-1}>\gamma$, we declare that the change happens at time $t$, otherwise, we proceed to the next time step. In an offline setting, the change point can be estimated as the time index $v$ where the changes of the average score is the largest:
\begin{flalign}\label{eq:changepoint}
v = \argmax_t (\bar{s}_t - \bar{s}_{t-1}).
\end{flalign}
After the change point is updated, we return again update the SDE model parameters and then the change point estimate again and repeat this process until convergence. We summarize our algorithm by pseudocode in Algorithm \ref{alg:cdgans}.

\begin{algorithm}[t]
\caption{Neural SDEs with Change Points}\label{alg:cdgans}
\begin{algorithmic}
\Require Initial parameters $\theta_0, \theta_1, \phi, \nu$, training samples $X^1_{1:T}, \cdots, X^N_{1:T}$.

\While{not converged}

\State Update $\theta_0, \theta_1, \phi$ by running SGD based on $\nu$.
\State Compute $\bar{s}_t$ using \eqref{eq:averagescore} based on current $\phi$
\State Update $\nu$ according to \eqref{eq:changepoint}.
\EndWhile
\end{algorithmic}
\end{algorithm}

{\bf Extension to multiple change points:} Our algorithm can be easily adapted to the cases where there are multiple changes. Assume that there is only one change within a window with size $w$. We sort all $s_t-s_{t-1}$ in descending order and denote their time index as $\hat{\nu}_1, \hat{\nu}_2, \cdots$. The change point is first declared as $\hat{\nu}_1$. If $|\hat{\nu}_2-\hat{\nu}_1| \leq w$, we discard $\hat{\nu}_2$ and proceed to the following element until we find the $i$ such that $|\hat{\nu}_i-\hat{\nu}_1| > w$. Then, $\hat{\nu}_i$ will be another change point. More change points can be found by repeating this process.

\begin{figure*}[t]
\includegraphics[width=\linewidth]{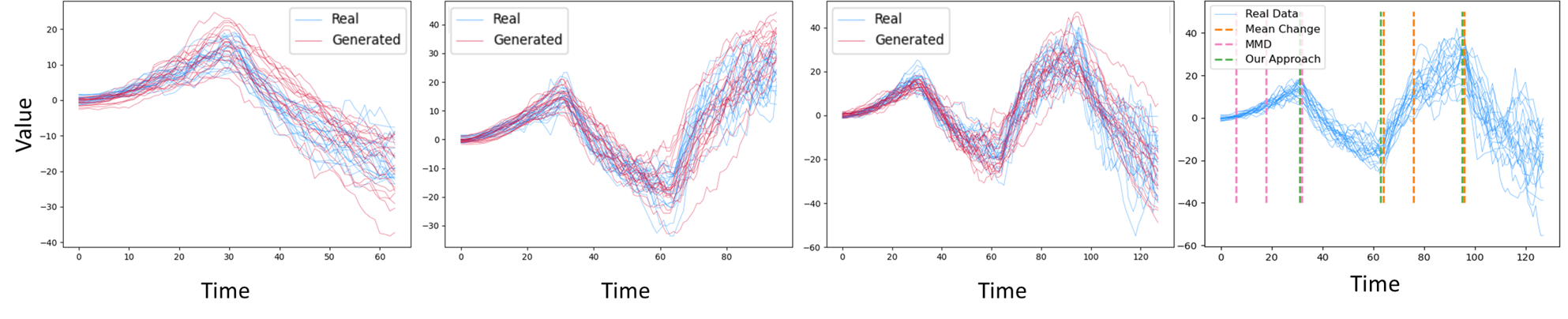}
\vspace{-0.5cm}
\caption{Simulation results on synthetic OU process data withe change points.}\label{Fig.1}
\end{figure*}

\section{Simulation Results}\label{sec:simulation}
\vspace{-0.08cm}
\subsection{Toy Example: Ornstein-Uhlenbeck Process}
\vspace{-0.08cm}
We begin by fitting a time-dependent one-dimensional Ornstein-Uhlenbeck (OU) process, which is defined by the following SDE:
\begin{flalign}
    dX_t  =(\mu t - \theta X_t)dt + \sigma\circ dW_t.
\end{flalign}
We consider the cases where there is one change point, two change points and three change points. Let the change points be $\nu_1 = 32, \nu_2 = 64, \nu_3 = 96$. Before $\nu_1$, we set $\mu_1 = 0.04, \theta_1 = 0.1, \sigma_1 = 0.4$. After $\nu_1$ and before $\nu_2$, we set $\mu_2 = -0.02, \theta_2 = 0.1, \sigma_2 = 0.4$. After $\nu_2$ and before $\nu_3$, we set $\mu_3 = 0.02, \theta_3 = 0.1, \sigma_3 = 0.4$. After $\nu_4$, we set $\mu_4 = -0.02, \theta_4 = 0.1, \sigma_4 = 0.4$.

{\bf Baselines:} We compare our approach with two heuristic change detection approaches. The first one detects the change by the mean change of the time series. Specifically, we partition the sample into different segments $X_{1:w}, X_{2:w+1}, \cdots, X_{T-w+1: T}$. Define the average mean of each segment over all training samples as
\begin{flalign}
\bar{\mu}_t = \frac{1}{N}\sum_{i=1}^N \sum_{t=1}^w X_{t}^{(i)}.
\end{flalign}
The change point using the average mean is then estimated as $\hat\nu_{\rm mean} = \argmax_{t}(\bar{\mu}_t-\bar{\mu}_{t-1})$.
% \begin{flalign}
% \nu = \argmax_{t}(\bar{\mu}_t-\bar{\mu}_{t-1}).
% \end{flalign}
The second approach is based on the maximum mean discrepancy (MMD) which is usually used to quantify the difference between two distributions. 
%Denote by $\text{MMD}(X, Y)$ the MMD between two %set of samples.
% $X = (X_1, \cdots, X_m), Y=(Y_1,\cdots, Y_n)$ is defined as follows:
% \begin{flalign}
% &\text{MMD}(X, Y) = \Big(\frac{1}{m(m-1)}\sum_{i\neq j}k(X_i,X_j)\nn\\& +\frac{1}{n(n-1)}\sum_{i\neq j} k(Y_i,Y_j) - \frac{2}{mn}\sum_{ij}k(X_i,Y_j)\Big)^{1/2},
% \end{flalign}
% where $k(\cdot, \cdot)$ is a kernel function. In this problem, we set $k(\cdot, \cdot)$ to be the Gaussian kernel with bandwidth 1. 
Define the average MMD between two consecutive segments as
\begin{flalign}
\bar{\eta}_t = \frac{1}{N}\sum_{i=1}^N \text{MMD}(X_{t-1:t+w-1}^i, X_{t:t+w}^i).
\end{flalign}
The change point using the average MMD is then defined as $\hat\nu_{\rm MMD} = \argmax_t \bar{\eta}_t$.
% \begin{flalign}
% \nu = \argmax_t \bar{\eta}_t.
% \end{flalign}

{\bf Results:} We first plot the generated sample paths of our approach and the training data in Fig.~\ref{Fig.1} for all three cases. It can be seen that our approach detects the change points and fits the training data well even when there are multiple change points. To compare our approach with the heuristic approaches, we plot the estimated change points for all approaches along with the training data for the case with three change points in the last figure of Fig.~\ref{Fig.1}. It can be seen that MMD and mean change don't reflect the change of the SDE trajectories while our approach detects the change accurately.

% \begin{figure}[htb]
% 	\centering 
% 	\subfigure[One Change Point]{
% 		\label{fig:1}
% 		\includegraphics[width=0.48\linewidth]{OU_one_change2.png}}
% 	\subfigure[Two Change Points]{
% 		\label{fig:2}
% 		\includegraphics[width=0.48\linewidth]{OU_two_change.png}}
%         \subfigure[Three Change Points]{
% 		\label{fig:3}
% 		\includegraphics[width=0.48\linewidth]{OU_three_change2.png}}
% 	\caption{Simulation Results on Synthetic Data}
% 	\label{Fig.1}
% \end{figure}

% \begin{figure}[ht]
% \includegraphics[width=3.2in]{comparison_change.png}
% \vspace{.3in}
% \caption{Comparison of Three Algorithms}\label{fig:heuristic_gans}
% \end{figure}

\subsection{Real Data Experiment: ETF Data}
We use part of the Exchange-Traded Fund (ETF) data from December 12, 2019 to June 07, 2020 which covers the COVID period where a sharp distributional shift occurred. Each sample of the data corresponds to have a different underlier of the S\&P 500 index. The data is normalized to have mean zero and unit variance. 

{\bf Baselines and metrics:} We compare our approach against two baselines: GAN-based neural SDEs without change detection (denoted by SDEGAN) \cite{kidger2021neural} and the RTSGAN \cite{pei2021towards}. For our approach, we consider the cases with one, two and three change points (denoted by CP-SDEGAN$^1$, CP-SDEGAN$^2$, CP-SDEGAN$^3$). We use three metrics to compare their performance. 

\textit{MMD}: We use MMD to measure the difference between the training samples and generated samples. Smaller value means the generated samples are closer to the training samples.

\textit{Prediction}: We perform one-step prediction under the train-on-synthetic-test-on-real (TSTR) metrics \cite{esteban2017real}. We train a 2-layer LSTM predictor on the generated samples and test its performance on the real data. Smaller loss means that the generated samples are able to capture the temporal dynamics of the training samples.

\textit{Classification}: We train a 2-layer LSTM to distinguish between the real data and the generated samples and get the classification loss on the test set. Larger loss means it's more difficult to distinguish the real and synthetic data.

% \begin{table}[t]
% \centering
% \resizebox{0.9\columnwidth}{!}{\begin{tabular}{ | m{3.5em} | m{1.7cm}| m{2cm} | m{2cm}|} 
%   \hline
%   Metric & MMD & Classification &Prediction\\ 
%   \hline
%   RTSGAN & 0.2942$\pm$6e-8 & 0.0680$\pm$0.0774 &1.0416$\pm$0.0005\\ 
%   \hline
%   SDE & 0.6028$\pm$1e-7 & 0.1716$\pm$0.0714 & 0.8189$\pm$0.0001\\ 
%   \hline
%   SDE1 & 0.2144$\pm$4e-7 & 0.2038$\pm$0.0682 &0.8316$\pm$0.0002\\ 
%   \hline
%   SDE2 & 0.1548$\pm$2e-8 & 0.1847$\pm$0.0351 &0.8173$\pm$0.0001\\ 
%   \hline
%   SDE3 & \textbf{0.1464$\pm$2e-8} & \textbf{0.2816$\pm$0.0867} &\textbf{0.8167$\pm$0.0002}\\ 
%   \hline
% \end{tabular}}
% \vspace{-0.2cm}
% \caption{Results for ETF data.}
% \label{table:1}
% \end{table}

\begin{table}[t]
\centering
\resizebox{\columnwidth}{!}{
\begin{tabular}{cccc} 
  \toprule
  & \textbf{MMD $\downarrow$} & \textbf{Classification $\uparrow$} & \textbf{Prediction $\downarrow$}\\ 
  \midrule
  RTSGAN & $0.2942 \pm 0.0000$ & $0.0680 \pm 0.0774$ & $1.0416 \pm 0.0005$\\ 
  SDEGAN  & $0.6028 \pm 0.0000$ & $0.1716 \pm 0.0714$ & $0.8189 \pm 0.0001$\\ 
  CP-SDEGAN$^1$ & $0.2144 \pm 0.0000$ & $0.2038 \pm 0.0682$ & $0.8316 \pm 0.0002$\\ 
  CP-SDEGAN$^2$ & $0.1548 \pm 0.0000$ & $0.1847 \pm 0.0351$ & $0.8173 \pm 0.0001$\\ 
  CP-SDEGAN$^3$ & \textbf{0.1464 $\pm$ 0.0000} & \textbf{0.2816 $\pm$ 0.0867} & \textbf{0.8167 $\pm$ 0.0002}\\ 
  \bottomrule
\end{tabular}}
\vspace{-0.2cm}
\caption{Results for ETF data.}
\label{table:1}
\end{table}

{\bf Results:} We summarize our results in Table \ref{table:1}. For this dataset, we have that our approach outperforms the RTSGAN \cite{pei2021towards} and neural SDEs without change detection \cite{kidger2021neural}. However, assuming different number of change points will lead to different performance of our algorithm on all three metrics. In real-world applications, based on specific tasks, we can determine the best number of change points to train neural SDEs by model selection.

\section{Conclusion}\label{sec:conclusion}
In this paper, we proposed a novel approach to detect the change of the neural SDEs based on GANs and further model time series with multiple SDEs conditioning on the change point. Our research contributes to the advancement of more robust and accurate modeling techniques, particularly in the context of financial markets, where the ability to capture dynamic changes is crucial for informed decision-making. Our results show that the proposed approach outperforms other deep generative models in terms of generative quality on datasets exhibiting distributional shifts.

\section{Acknowledgements}
This paper was prepared for informational purposes by the Artificial Intelligence Research group of JPMorgan Chase \& Co. and its affiliates (``JP Morgan''), and is not a product of the Research Department of JP Morgan. JP Morgan makes no representation and warranty whatsoever and disclaims all liability, for the completeness, accuracy or reliability of the information contained herein. This document is not intended as investment research or investment advice, or a recommendation, offer or solicitation for the purchase or sale of any security, financial instrument, financial product or service, or to be used in any way for evaluating the merits of participating in any transaction, and shall not constitute a solicitation under any jurisdiction or to any person, if such solicitation under such jurisdiction or to such person would be unlawful.

\bibliographystyle{IEEEtran}
\bibliography{ref}

\end{document}